\RequirePackage{etoolbox}
\csdef{input@path}%
{%
 {sty/},
 {img/},
}
\documentclass{elsevierbook}

\usepackage{natbib}
\usepackage{url}

\begin{document}


\Mainmatter

  \cleardoublepage

\begin{frontmatter}
\chapter{Applications and Challenges of SA in Real-life Scenarios}\label{chap2}

\begin{aug}
\author[addressrefs={ad1}]%
  {\fnm{Diptesh}   \snm{Kanojia}}%
\author[addressrefs={ad2}]%
 {\fnm{Aditya} \snm{Joshi}}%
\address[id=ad1]%
  {Surrey Institute for People-centred AI, United Kingdom}%
\address[id=ad2]%
  {SEEK, Australia}%
\end{aug}


%



  


\end{frontmatter}

\begin{abstract}
    Sentiment analysis has benefited from the availability of lexicons and benchmark datasets created over decades of research. However, its applications to the real world are a driving force for research in SA. This chapter describes some of these applications and related challenges in real-life scenarios. In this chapter, we focus on five applications of SA: health, social policy, e-commerce, digital humanities and other areas of NLP. This chapter is intended to equip an NLP researcher with the `what', `why' and `how' of applications of SA: what is the application about, why it is important and challenging and how current research in SA deals with the application. We note that, while the use of deep learning techniques is a popular paradigm that spans these applications, challenges around privacy and selection bias of datasets is a recurring theme across several applications.
\end{abstract}
The previous chapter describes sentiment (SA) and its allied sub-problems via the perspective of computational intelligence. At its heart, SA is a prediction problem over text. Given a piece of text, the task is to predict sentiment (positive/negative/neutral) or emotion (happy/sad/angry and so on), \textit{etc.} Aspect-based SA, domain-specific SA can be viewed as sub-tasks in SA- each bringing additional challenges providing opportunities for innovative architectures. 

However, any area of technology is centred around two pillars: innovation and usefulness. The two are interleaved in that one drives the other. The case is no different for SA. The innovation and advancements in SA can be attributed to the availability of benchmark datasets, most notably for English as a part of the GLUE benchmark~\cite{wang2018glue} or Sentiment Treebank. SA witnessed a development of large-scale datasets and lexicons. For example, Amazon made their review dataset public~\citep{ni-etal-2019-justifying} providing an impetus to research in aspect-based SA. Sentiment tree bank provided a tree-based dataset of sentences and phrases labelled for sentiment. As a result, SA attracted attention~\cite{socher2013recursive}. There were datasets to train large-scale neural models, and the public availability of the datasets meant reproducibility and comparison of proposed approaches. 

While availability of benchmark datasets spawned innovation in SA, this chapter is about the second pillar: usefulness. We visualise usefulness of SA in the form of its last mile applications. The phrase `last mile' is often used in the context of e-commerce websites: websites that sell products to consumers. The Cambridge dictionary describes `last mile' as the last stage in a process, especially of a customer buying goods. In the context of SA, the last mile are the applications of SA to the real world. The usefulness of SA is manifested in the real-world applications of SA: from helping stock traders make purchase decisions to mining continuous streams of social media data for intelligence on public health outbreaks. In this chapter, we cover these applications of SA in real-life scenarios: health, culture, social policy, human safety and other areas of NLP research. 

\begin{figure}
    \centering
    \includegraphics[width=\textwidth]{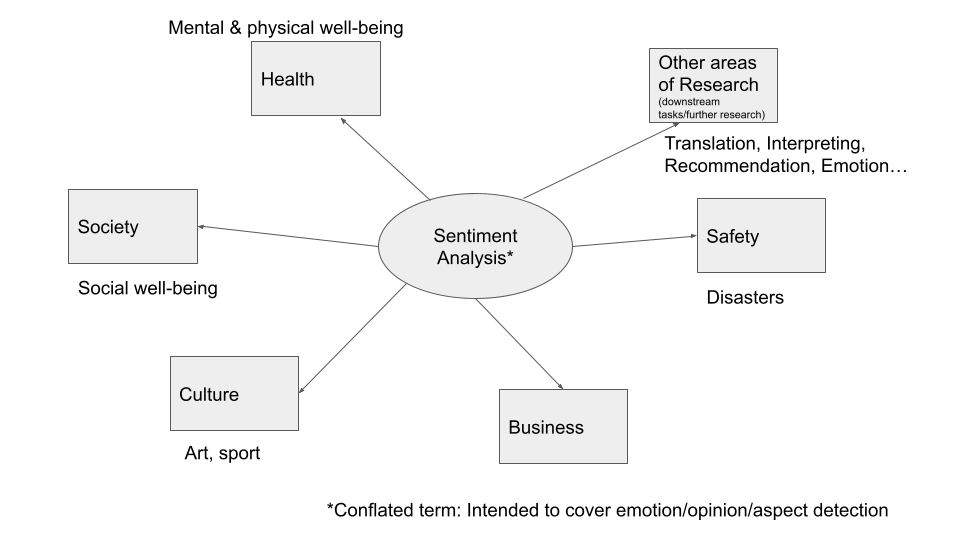}
    \caption{Applications of SA}
    \label{fig:overview}
\end{figure}

Figure ~\ref{fig:overview} is a perspective of what constitutes as an application of SA. SA is at the centre of the figure. It is conflated to mean SA, emotion analysis, aspect-based SA and other sub-areas of SA. The focus of this chapter is, however, the connections of SA with these other areas that are positioned around it: society, health, safety, culture, business and other areas of NLP\footnote{This is not the complete list of applications of SA. Few applications such as education technology or disaster management are not included in the chapter.  However, we believe that the applications covered in the chapter are representative of majority of the applications, and provide a sense of the utility of SA.}. Each of these form forthcoming sections of he chapter. In each section, we address questions pertinent to the specific domain. For example, how can SA be applied to improve health of individuals and community? How can SA be applied to aid businesses achieve their objectives? The areas discussed in the paper are real life scenarios: scenarios impacting individuals, communities or organisations. 

At the outset, it is important to note that an application does not only mean the domain of the dataset used. A new application domain may provide unique opportunities to improve SA. For example, ontologies created by medical experts are unique to the health domain and have shown to be immensely useful as structured knowledge bases for SA applications in health. Therefore, in addition to applications of SA, we also highlight peculiar challenges and opportunities for these applications to demonstrate that an effective application often needs application-based understanding and adaptation of SA.

In each of the following sections, we describe applications and challenges of SA in real-world scenarios, as illustrated in the Figure above.

\section{Health}\label{sec2.1}
Advances in science and technology have been applied and deployed to improve health and well-being of individuals and the society have always received attention. SA is no exception. Health-related textual datasets may be generated by medical professionals in the form of clinical notes or by individuals in the form of social media posts. SA applied to each of these can be useful in many ways. In this section, we now discuss some of these applications.
\subsection{Challenges}
The immediately apparent difficulty in applying SA to health-related problems is the availability of datasets. Medical datasets may not always be available in digitised form in some regions of the world. Similarly, privacy and confidentiality of medical data prevents making it accessible for research or deployment. However, some benchmark datasets such as MIMIC are available for research purposes. An alternative to the medical datasets are user-generated datasets such as tweets. Since official medical datasets are confidential, tweets posted by people can be a useful alternative. The epidemiology community, therefore, refers to tweets as `open-source data'. 

However, using tweets for healthcare applications of SA has its own challenges. Social media usage is popular in certain demographics, potentially based on their age or acquaintance with technology. As a result, as a volume, social media-based datasets may not be representative of a community, reflecting a selection bias.
\subsection{Applications}
In situations where medical transcripts and data by medical practitioners is available, \citet{denecke2015sentiment} identify the following avenues for application of SA using medical transcripts:
\begin{enumerate}
    \item Change in health status: The medical practitioner may note whether a patient felt better or worse. This may be done using sentiment words. This sentiment may be expressed towards entities such as the pharmacological interventions used for treatment.
    \item Critical events: Medical emergencies can be reported through strong negative sentiment.
    \item Patient experience: A medical practitioner may report patient experience as the patient describes how they are feeling during an appointment.
\end{enumerate}

The points above hold true for user-generated medical data as well. Discussion forums on the internet for people suffering from certain medical conditions allow people to share experiences with each other. SA applied on these discussion forums can help to identify what measures are working for people - and what are not.

In addition, since social media is widely used, people often report their health conditions in the form of social media posts such as tweets. Social media-based epidemic intelligence is the use of social media posts to monitor initiation, prevalence and spread of epidemics in the real world~\citep{adityaepidemiccsur}. By its very definition, the word `disease' signifies a situation where a person is not (`dis') at `ease' with themselves. As a result, people experiencing illnesses typically express negative sentiment, at least implicitly negative. Social media-based epidemic intelligence (SMEI) is a quintessential application of SA applied to the world of health/healthcare.

At the heart of SMEI is the task of predicting whether a tweet mentioning an illness-related word is a report of the illness. Such a tweet has been called a personal health mention. For example, a tweet `This fever is killing me' is a report of an illness, while `Fever, cough and loss of taste are common symptoms of COVID-19' is not a report of an illness although both contain the symptom word `fever'. The task of computationally detecting personal health mentions is called personal health mention classification. Personal health mention classification uses features based on sentiment, as given in ~\citet{adityaepidemiccsur}: the number of negative words, the implied sentiment and so on. Alternatively, sentiment information can be incorporated into a neural architecture. \citet{biddle2020leveraging} concatenate word representations with sentiment distributions obtained from three sentiment lexicons. This is then passed to sequential layers such as LSTM before the prediction is made.

\begin{figure}
    \centering
    \includegraphics[width=0.8\textwidth]{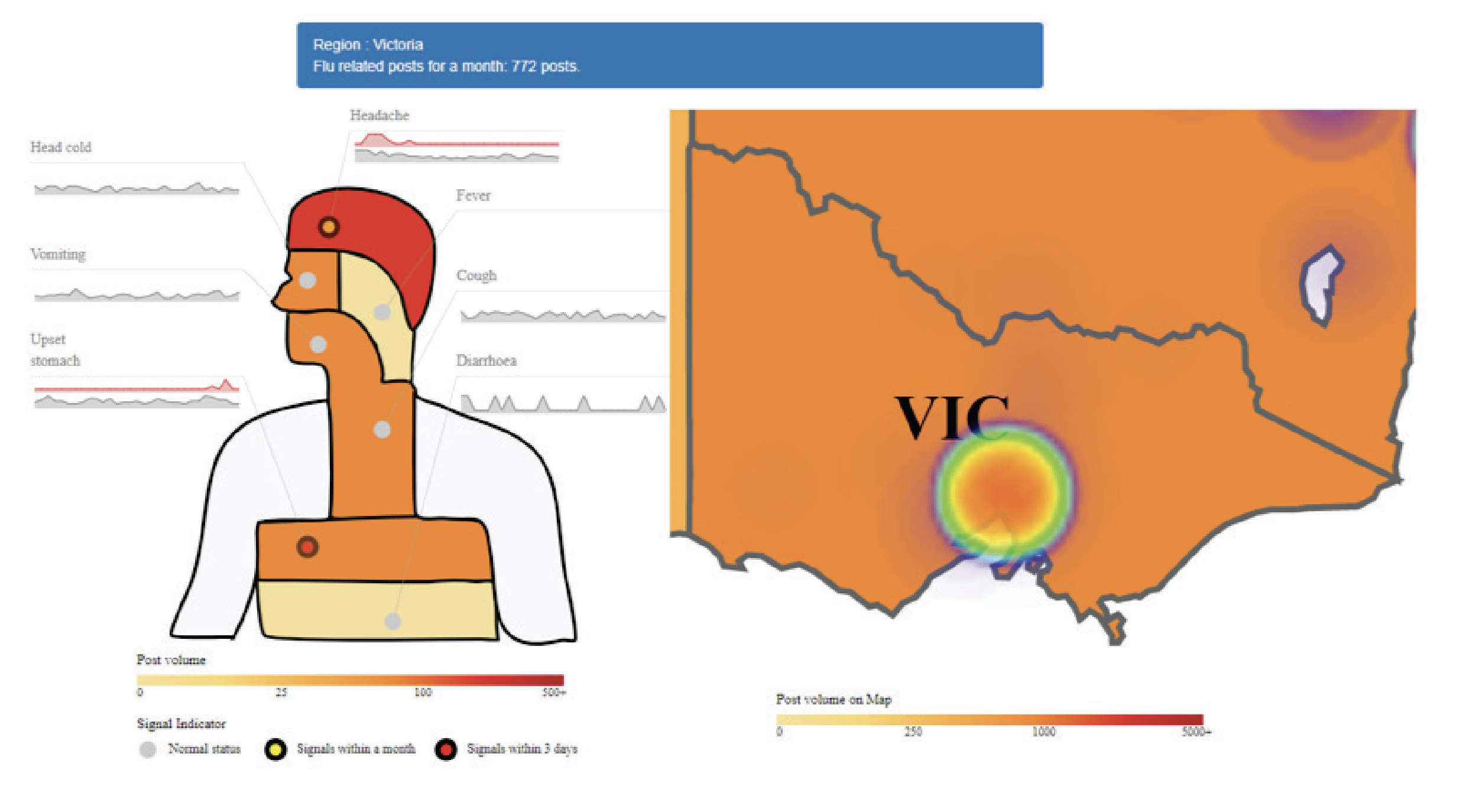}
    \caption{Social Media-based Epidemic Intelligence}
    \label{fig:wtflu}
\end{figure}

The second step in SMEI is health event detection. It uses time series monitoring in order to predict an outbreak in number of personal health mentions (say, within a time period). SMEI can potentially detect short-term outbreaks as well as beginning of long-term epidemics. SMEI can eventually be deployed in a dashboard that monitors social media posts in a geography of interest. One such dashboard is `Watch The Flu'\citep{jin2020watch} that automatically monitors tweets posted from within Australia, identifies the ones that are personal health mentions (via classification) for influenza symptoms and then detects potential spikes in symptoms reports (via time series monitoring). A screenshot from their paper is given in Figure~\ref{fig:wtflu}.

SA also assumes importance in terms of the aspects towards which the sentiment is expressed. For example, vaccination sentiment detection involves detection of sentiment towards vaccines~\citep{kummervold2021categorizing}. Similarly, Social Media Monitoring 4 Health (SMM4H) workshop conducts shared tasks for health-related tweets~\citep{magge-etal-2021-overview}. The reported systems regularly report the use of SA-based enhancements for tasks such as adverse drug reaction detection, vaccination behaviour detection and so on. 

\begin{figure}
    \centering
    \includegraphics[width=0.9\textwidth]{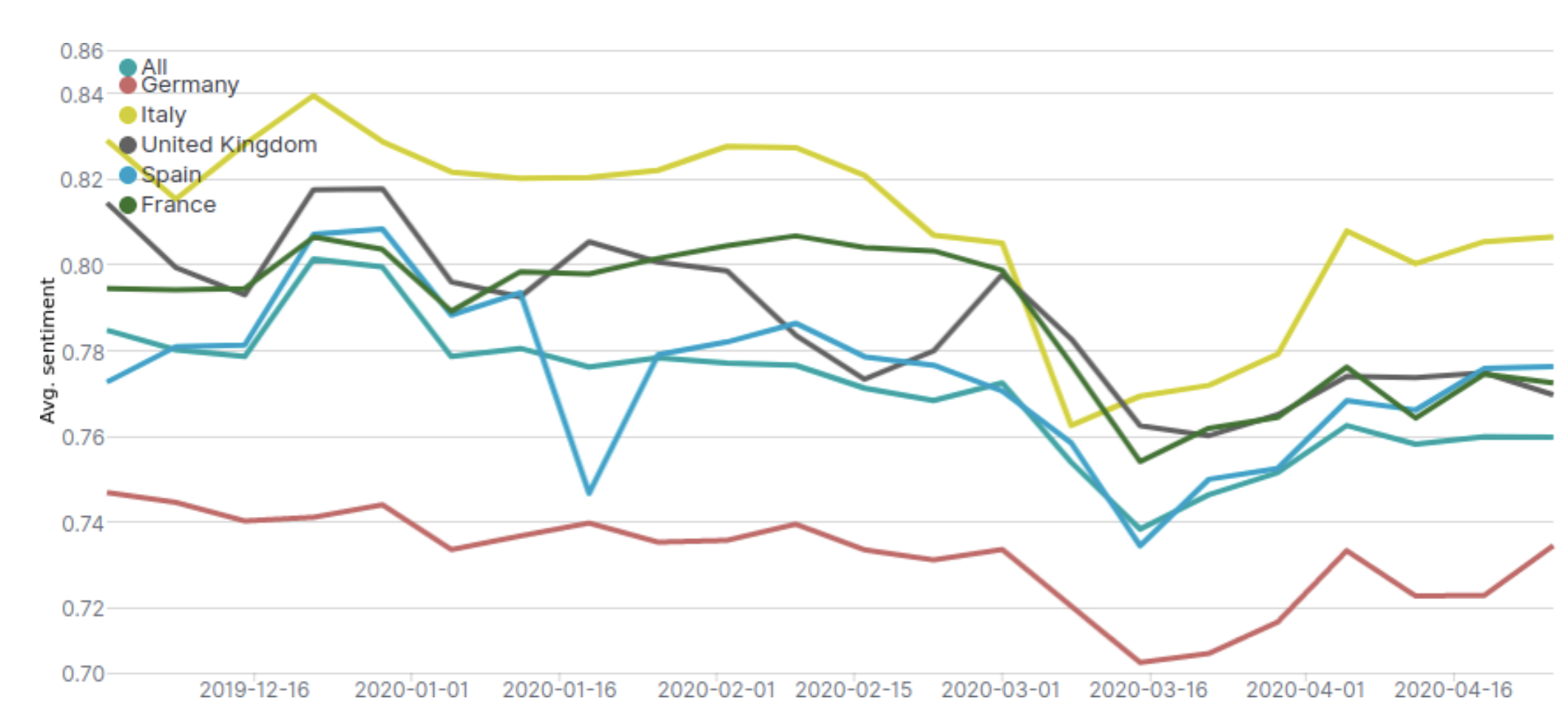}
    \caption{Sentiment changes in different countries during the COVID-19 pandemic}
    \label{fig:sentigraphcovid}
\end{figure}

The COVID-19 pandemic is an unprecedented pandemic that resulted in worldwide infections, deaths, lockdowns and economic slowdowns. In ~\citet{kruspe2020cross}, SA (using a neural network) is applied to multi-lingual tweets from around the world. The paper shows how sentiment towards the pandemic changed in different parts over the world over time. An application of SA such as this provides a zeitgeist of the impact of the COVID-19 pandemic. A figure from this paper is included as Figure~\ref{fig:sentigraphcovid}.

SA of tweets during the time of a pandemic can help to understand issues being experienced by people.  An example in the context of COVID-19 is ~\citet{kleinberg-etal-2020-measuring}. The paper reports a COVID-19 Real-world Worry Dataset. The datasets contains tweets posted from within UK during the time of the COVID-19 pandemic. In the introductory paper, the authors use topic modeling and lexical analysis using Linguistic Inquiry and Word Count (LIWC) in order to understand issues and concerns as expressed in the tweets.

\section{Social Policy}\label{sec2.2}

SA holds an important place in the social conversation space as combined with text mining from the social media domain. It can be a really powerful tool to gather public opinion. Early research in this area~\citep{karamibekr2012sentiment} argues that this is an interdisciplinary field that crosses the boundaries of artificial intelligence and natural language processing- the domains to which it is known to belong to. As the digitisation of public opinion via forums like Reddit and Twitter are becoming more popular, with more and more people getting access to the internet, we see a shift in the strategies employed by governments of the world that take these digital media into account. The analysis of sentiment towards political discourse is also important as sentiment/emotion expressed towards a political scenario are polar in nature. People express opinions in favour or against a political decision or even the rulings of courts in many countries. Therefore, it is widely believed and also already acceptable to use SA-based tools and research for measuring the sentiment of people towards social policy. A social issue is an issue that relates to people's personal lives and interactions. The impact of public opinion on social issues and on policy makers operates in a similar manner to those of product reviews for manufacturers. Current governments and organisations engaging with social issues can analyse public criticisms or supports for a particular policy and consider public opinions in making decisions. Politicians can receive electorates' opinions concerning important issues and their expectations. However, social issues are very different from products and services. The divide on social issues can be much larger in people compared to the divide in opinion on a product. Such opinions can also be very nuanced and may be expressed neither in complete favour nor completely against a social policy, which is also a challenge faced by SA research when applied to the social policy domain. People who voice their opinions on social media platforms also tend to use \textit{hashtags}, \textit{slangs}, \textit{mixed language}, \textit{i.e.,} code-mixing/switching, and inconsistent \textit{variants of words} in a language which are technical challenges to further mining the social opinion of a populace. There are conversations in comments and on Twitter threads that imbibe a context for humans, whereas this context combined with the background knowledge of the subject matter is impossible for machines to comprehend and take into account, which is mining social opinions. Despite these challenges, the applications of SA research to the social policy domain continues to thrive and has allowed so many research questions to stem from this field which we will discuss further in the next subsection. 

\subsection{Applications in Social Policy}

Social media provides opportunities for policy makers to gauge public opinion. However, the large volume and a variety of expressions on social media have changed traditional policy analysis and public sentiment assessment. The applications of SA research in this area has an immense contribution in gauging the social and political sentiment of a populace. For example,~\citet{doi:10.1177/1532673X20920263} show how public opinion shifted post-legalisation of same-sex marriage in the United States. They use Twitter to analyse public opinion after the Supreme court ruling and observe how states affected by the ruling carried a more negative sentiment towards the subject. Court rulings on social policy can polarise the public as the mandate might not be culturally acceptable by the layman.~\citet{crossley2017sentiment} propose a tool they call SEANCE for sentiment, social cognition and social-order analysis. This tool offers various features for the analysis of data collected from various social media forums. Existing research has also presented arguments that political scientists have done little in the way of rigorous analysis towards the sentiment/mood of the general public~\citep{durr1993moves}. However, SA research, due to automated computational approaches, is able to do this without the manually-conducted political surveys and opinion polls. With the rapid growth of social media content and usage, policy makers and even ordinary citizens are able to efficiently obtain voluminous data about public sentiment. Opinion leaders, authorities, and activists who share their ideas on Twitter are often followed closely by thousands of users. These leaders provide valuable content as well as linkage information that can offer insights for policy decision making~\citep{chung2016social}. It is well documented that online forums like Twitter affect the political discourse and raises questions that need answers from the governments~\citep{parmelee2011politics}. These questions can be related to incidents that happen rarely or can be related to the long-term political decisions or a potential problem for the country in the long-term ( ageing population~\citep{petonito2015silver} ). SA research has focused on public opinion about rural issues as well- issues pertaining to a section of the population that has limited to the internet where these issues are discussed, which is ironic to think about~\citep{vagrani2020appraising,botterill2009role,singh2018sentiment,singh2019sentiment,li2021sentiment}. Social or political campaigns run by the governments/politicians are also topics of discussion amongst the online population which can help predict the public opinion towards a campaign~\citep{tayal2017sentiment,sandoval2020sentiment,hoffmann2018too,budiharto2018prediction,chen2021measuring}. The analysis of online text or opinion can help create a positive impact on the policy as governments can be better informed about public opinion. Such content can be useful for government agencies, as it can help them understand the needs and problems of society to formulate effective public policies. The constantly increasing complexity of social problems and needs makes this political content even more valuable, as it contains extensive knowledge concerning such problems, which can be quite useful for understanding and managing their complexity~\citep{charalabidis2015opinion}. Online media can affect and help form social policies, as evident by these significant number of studies in this domain. However, the mining of sentiment/opinion from public forums does not come without its own set of challenges which we discuss in brief in the next subsection.

\subsection{Challenges in Social Policy}

Social media is full of opinions, emotions, and sentiments expressed largely in the forms of text. This voluminous amount of text, however, can be expressed in different ways, which is largely considered a challenge by the researchers in the SA domain~\citep{tsou2015research}. A whole subspace of SA researchers have dedicated themselves to identifying genuine users and account holders as compared to what are known as fake or `troll' accounts online. Fake accounts on social media are run via politically motivated organisations with the help of a large number of employees and also by `bots' which can help further any agenda. Here are some of the research challenges faced by SA researchers when collecting the data from social media forums:
\begin{description}
    \item[Identification of Users:] While traditional data collection methods like surveying and interviewing largely relied upon manual collection of data, it could collect authentic data. However, with the advent and easy access to social media, it is an important challenge to find out which accounts carry the opinions of genuine social media users.~\citet{Flores-Saviaga_Keegan_Savage_2018} note that political trolls initiate online discord not only for the laughs but also for ideological reasons. They uncover that the most active users employed distinct discursive strategies to mobilise participation and also deployed technical tools like bots to create a shared identity and sustain engagement. In fact, they themselves face a challenge as they build their own `troll' community while simultaneously disrupting others.
    \item[User Privacy:] Privacy concerns and related risks are a cause of concern for researchers and for the public. The danger of spatial information disclosure becomes more serious when people use mobile devices and ``check-in'' through social media to reveal their physical locations. Although a large-scale opinion mining research practice may ignore such data, released datasets may contain sensitive information, and hence, researchers face an additional challenge in terms of anonymisation of data streams used for research. This information can at times also reveal medical records and public health data, which should not be shared without ensuring either- the consent of the patient or proper anonymisation techniques. 
    \item[Noisy Data:] Rapid growth of social media has also led to a rapid increase in spam and online marketing of content which is prevalent in social conversations. This noise may create a hurdle for SA research that tries to capture a genuine opinion of social media users. Non-relevant posts and comments, marketing messages, bots, `\textit{clickbaits}' and advertisements lead to a lot of noisy data, which needs to be removed before the analysis of user data generated via social media forums.
    \item[Mental Health:] Mental illness is a serious and widespread health challenge in today's day and age. Many people suffer from depression, and only a fraction opt to receive adequate treatment. The role of this need for validation via social media posts has been investigated by many researchers~\citep{de2013role,berryman2018social}. Advocates of mental health and policy makers have expressed concerns regarding the potential negative impact of social media on an overall young population. Although much of the public narrative on such negative effects implies that its mere exposure is related to common mental health problems, the evidence suggests that quality rather than quantity of use is more crucial. For example, existing research argues that the use of social media for negative social comparison, which, alongside rumination, leads to depression.
    \item[Language:] SA research is a part of the NLP domain and suffers from similar technical challenges in terms of the language used to express opinions on the internet. Automatically detecting the sentiment or opinion from online forums is challenging in terms of the language used on social media. For example, emotions are not stated explicitly with the content. Social media text can express frustration, depression, anger, joy, sarcasm and so on in varying degrees which are implicitly captured in the text. Certain words which carry negation towards the sentence can be very impactful towards the sentiment but are not used to capture the sentence polarity. The use of creative and non-standard language, which includes hashtags and abbreviations, popular among a young generation can also convey sentiment and emotion which need to be captured. Para-linguistic information like facial expressions is often ignored as SA research primarily deals with text, but this information also carries sentiment which should be captured. Cross-cultural differences in emotions expressed towards a particular situation can also lead to ambiguity which sometimes, even humans can not comprehend. Of course, the lack of labelled sentiment data is also a huge challenge which leads to the use of semi-supervised and unsupervised approaches which can not be relied upon fully. 
\end{description}

The mining of opinion or sentiment from the social media text, especially when social policy is concerned, faces hurdles due to such challenges. As research in SA progresses, new sub-problems like aggression/hate-speech detection, offensive language identification, and emotion analysis have emerged, which attempt to tackle these in a very ``one-problem-at-a-time'' way. Hopefully, in the near future, the solutions to these challenges will also emerge slowly and help us garner actual public opinion about social policies from the internet.

\section{E-Commerce / Industry}\label{sec2.3}

Customer reviews over online e-commerce platforms have become a common source of product feedback for consumers and manufacturers alike. Similarly, services like Yelp (\url{www.yelp.com}), Zomato (\url{www.zomato.com}), and TripAdvisor (\url{www.tripadvisor.com}) are commonly available for customer feedback on restaurants and travel destinations. Social media platforms like Facebook and Instagram are also known hosts to pages and channels which provide feedback on restaurants and travel destinations. These review sources also act as an important part of the `\textit{feedback loop}' for e-commerce platforms and manufacturers to improve the overall product quality based on certain aspect-level reviews. Recent research shows that fine-grained aspect-based SA has penetrated SA research in NLP~\citep{fan-etal-2018-multi,angelidis-lapata-2018-summarizing,schmitt-etal-2018-joint,hoang-etal-2019-aspect,jiang-etal-2021-attention}.
In this section, we discuss how SA has gained importance in the e-commerce domain and also provide an outline of the challenges posed for this research.

\subsection{Applications in E-Commerce}

Customers who buy products online often rely on other customers' reviews found over e-commerce platforms, more than reviews found in specialist magazines. These reviews often prove to be important recommendations for books, mobiles, laptops, clothing \textit{etc.}, and seem reliable as they are written by product users, in most of the cases. In fact, short reviews form a separate category of their own and are known as `tips'. For example, ``\textit{I love this phone! The battery life and camera quality have won me over.}'' can be considered a review whereas ``\textit{Great battery. Nice camera. Photos are good}'' is considered a `tip' or short recommendation~\citep{ni-etal-2019-justifying}. The significantly increased exposure to digital technology and the penetration of the Internet in rural areas in recent years has led to a vast amount of expressed opinions over products. In particular, customers express opinions on several aspects of products, services, blogs, and comments which are deemed to be influential, especially when making purchase decisions based on product reviews~\citep{schouten2015survey}. This is in contrast with the traditional surveys and questionnaires that participants had to fill without any personal motivation, thus resulting in the sub-optimal collected information.

Many potential consumers read these opinionated reviews resulting in an influenced buying behaviour~\citep{bickart2001internet}. From the point of view of a consumer, the information provided by past consumers of a product is regarded as more trustworthy than the information provided by its manufacturer. Also, from a manufacturer point of view, 1) understanding the product performance in a real-world setting, 2) improving some aspects of existing products, and furthermore, 3) understanding how the information in product reviews interacts with each aspect of the product or the service, enables the manufacturer to take advantages of these reviews and improve sales. 

Recent NLP research on aspect-based SA (ABSA) has been driving the efforts of the e-commerce industry to analyse product reviews and gain the insights discussed above. While performing aspect-based SA, three distinguishable processing steps can be 1) identification of aspects and sentiment expressed, 2) classification of sentiment for each aspect, and 3) aggregation of the overall sentiment in the review. However, some studies only focus on one of them and do not perform all three of these steps. \textbf{The first step} involves the detection of aspects from the review text, which can be performed with the help of different approaches \textit{viz.} frequency-based, syntax-based, supervised machine learning-based, unsupervised machine learning-based, and hybrid approaches. \textbf{The second step} which entails the classification of sentiment, can also be performed with the help of various approaches, \textit{viz.} dictionary-based, supervised machine learning-based and unsupervised machine learning-based approaches. Additionally, there are methods that perform the detection of aspects and the analysis of sentiment using a joint method via different approaches such as syntax-based, supervised machine learning-based, unsupervised machine learning-based and hybrid approaches. We will discuss each of these approaches in brief here.

~\citet{hu2004miningAAAI, hu2004miningSIGKDD} observe that reviews contain a limited set of vocabulary, which is used more often, as likely aspects, in the reviews. Such frequent words inspired a straightforward approach for aspect detection based on nouns and compound nouns in the review. However, such an approach has clear shortcomings in terms of the existence of other nouns in the review, which are frequently used but are not aspects. Similarly, there are implicit aspects in the reviews that can not be detected via such an approach.~\citet{hai2011implicit} propose an approach to detect such implicit aspects with the help of association rule mining. They restrict sentiment words in the text to appear as rule antecedents only, and aspect words to appear as rule consequents, and generate association rules to find aspects based on pre-identified sentiment words. 

Syntax-based methods, on the other hand, focus on syntactical relations to find aspects present in a review text. One of the simplest approaches is to tag the text with part-of-speech categories for each token in the review and then identify the adjectives which modify a potential `aspect'. For example, the phrase `lovely ambience' where the `ambience' of a location is the aspect.~\citet{zhao2010generalizing} discuss how syntactic patterns in the text could be utilised to extract aspects using a tree kernel function. Similarly,~\citet{qiu2009expanding}, propose a method for aspect extraction which uses a double propagation algorithm for sentiment word expansion and aspect detection (later extended by~\citet{zhang2010extracting} and~\citet{qiu2011opinion}). This method treats aspect detection and sentiment lexicon expansion as interrelated problems and features the use of sentiment lexicon to find more aspects, and with additional aspects, this method is able to further find more sentiment lexicon.

Among the machine learning-based methods to detect aspects in a text, a limited number of supervised methods have been explored. Aspect detection can be modelled as a sequence labelling problem but supervised learning heavily relies on features extracted from the text; features that can provide more information than simple bag-of-words or part-of-speech tagged tokens in the text. However, a Conditional Random Field (CRF) -based method has been utilised to perform this task~\citep{jakob2010extracting}. This method utilises multiple features from the text, including the tokens, their part-of-speech tags, dependency relations between aspect token and sentiment lexicon, and the context of the aspect token. However, as compared to supervised learning-based methods, unsupervised learning-based methods have been more popular for the task of aspect detection. Most of the unsupervised methods utilise some variation of Latent Dirichlet Association (LDA) to explore the review text~\citep{blei2003latent}. 

The use of LDA to detect fine-grained aspects in a review text is not straightforward, as it is more suited to work with large documents which have multiple topics. LDA utilises a bag-of-words approach at the document level and hence end up detecting global topics, which is not very useful for detecting aspects. Therefore, for finding local aspects to a particular review, researchers combine LDA with Hidden Markov Model (HMM) to distinguish between aspect tokens and other background lexicon~\citep{lakkaraju2011exploiting}. An extension to LDA, known as Multi-grain LDA (MG-LDA), also attempts to solve the global \textit{vs.} local topics by having a fixed set of global topics and a dynamic set of local topics. For the local topics, the document is modelled as a set of sliding windows. This allows for a window overlap to sample one topic from multiple windows and the inherent bag-of-words modelling in LDA to have a larger set of words~\citep{blei2001topic,mei2007topic}. Then, there are the other variants that try to apply LDA at the sentence level~\citep{lu2011multi}, estimating the emphasis of aspects~\citep{wang2011latent}, adding syntactic dependencies~\citep{zhan2011semantic}, aspect taxonomy construction based on syntactic rules~\citep{luo2018extra} and incorporating product categories to alleviate the `\textbf{cold-start}' problem caused due to a limited number of reviews~\citep{moghaddam2013flda}. These efforts towards mining fine-grained aspects from reviews on e-commerce platforms show how research in SA is directly applicable in a real-world setting. At this point in the chapter, we encourage the reader to find out more about hybrid methods for aspect mining from reviews. These methods combine multiple approaches like serial hybridization~\citep{popescu2007extracting,yu2011aspect,raju2009unsupervised}, and parallel hybridization~\citep{blair2008building} to mine aspects. We shall now move on to discussing the second part of ABSA, which is SA/classification.

\textbf{SA}, in the context of this discussion, becomes the task of assigning a sentiment label or a score to each aspect in the review text. On E-Commerce platforms, each review text may contain multiple aspects or, in some cases, none of them. Therefore, SA research applied to such reviews should be robust to be able to handle any corner cases. The earlier methods for this task attempted to utilise dictionaries like Wordnets to find adjectives and label them as sentiment class (\textit{i.e.,} positive or negative). There are various heuristics among the early dictionary-based methods where the distance from a sentiment word would determine the appropriate sentiment class for the sentence~\citep{moghaddam2010opinion,hu2004mining,zhu2009multi}. However, as compared to aspect detection, sentiment classification can be easily modelled as a supervised learning task due to the reduction in the number of labels. Hence, we discuss the supervised machine- and deep learning-based methods briefly below.

\textbf{Supervised methods} for machine and deep learning have been significantly more successful for SA in the e-commerce domain. These methods can easily incorporate information from the tokens in the review text as features, and raw scores based on the sentiment lexicon can be used to compute additional feature values.~\citet{blair2008building} employ such a method and a MaxEnt classifier to obtain sentiment scores. Similarly, SVM is used by~\citet{yu2011aspect} to classify text into positive or negative labels by using a parse tree and reasonably assuming that each sentiment word found in the text should be within a distance of five `parse-steps' of an aspect.~\citet{choi2008learning} perform SA on very short expressions and associate these short expressions to aspects in the text. A lot of researchers use bag-of-words as features for supervised sentiment classification, but many argue that such models are too simple to be applied to a domain as vast as e-commerce. We highlight the challenges of using such approaches in the e-commerce domain in the next subsection and also discuss some solutions proposed. However, as opposed to the consistently tested classification-based methods, various researchers have attempted at using a regression-based method for obtaining sentiment scores. This does allow the sentiment score generation to be a little more fine-grained, but the eventually authors try to club the sentiment scores in five classes akin to the five start rating system.~\citet{lu2011multi} use a support vector regression model to obtain a sentiment score for each aspect. This model allows the sentiment score to be modelled as a real number in the zero to five rating interval. Similarly, with the help of PRanking~\citep{crammer2001pranking}, a perceptron-based online learning method,~\citet{titov2008modeling} perform SA given the topic clusters using an LDA-like model. The input to this method is different n-grams along with binary features which describe LDA clusters. This method is interesting as it involves the construction of a feature vector for each sentence based on the absence or presence of a certain word-topic-probability combination and then groups these probabilities into \textit{`buckets'}. The PRanking algorithm takes the feature vector and learned weights as an input to arrive at a number, which is checked against the boundary value to be divided into one of the five classes for a sentiment score. 

\textbf{Unsupervised methods}, however, are not significantly used for sentiment classification. With the help of parsed syntactic dependencies, one can find a potential sentiment phrase in the vicinity of a pre-detected aspect. Such a sentiment word carrying a polarity can be used to arrive at a decision regarding the overall sentiment of the aspect, \textit{viz.,} positive or negative~\citep{popescu2007extracting}. The research in SA has also attempted to borrow from the area of computer vision, specifically, an unsupervised method known as relaxation labelling~\citep{hummel1983foundations}. The relaxation labelling technique helps assign a polarity label to each sentiment phrase while adhering to a set of specified constraints, which can be obtained based on the semantic orientation of adjectives~\citep{hatzivassiloglou1997predicting}. The final output for any unsupervised method is also the same, either a positive or a negative label for each review text. 

The methods discussed above provide an insight into the research area of aspect-based SA and how they can be used in the e-commerce domain. Depending on the availability of data samples and available methods- SA research has been prevalent in shaping the online platforms which are now a part of our daily lives. Improving customer experience and building product brands are two very simple use-cases of this research in a real-world scenario. E-commerce giants also use SA to gain a competitive edge and engage with customers based on their SA research. However, no area of research or research methodologies exist without challenges of their own. In the upcoming section, we discuss these challenges in detail and show how SA research has tried to tackle some of them.

\subsection{Challenges in E-Commerce}

The premise for the SA research carried out by NLP researchers and e-commerce industries is the trustworthiness and reliability of these reviews posted by customers. However, the trust in these reviews is often misplaced due to `fake reviews'. For example, some authors glorify their own books in reviews on an e-commerce platform. This phenomenon is also known as \textbf{`sock-puppetry'}. Unfortunately, fake reviews can be produced in different ways such as,

\begin{itemize}
    \item product manufacturers may hire a group of individuals to post positive reviews of their products on online platforms,
    \item service providers like hoteliers and restaurant owners may glorify their services from multiple accounts/handles,
    \item authors who can post glorifying reviews about their work on online platforms via different handles/channels/social media influencers, and lastly,
    \item any of these parties can also hire people to post negatively about their competition's products and rate them lower on purpose.
\end{itemize}

Such fake reviews need to be identified and filtered before processing product reviews for the sentiment. One of the major challenges for SA research is the identification of these deceptive reviews. The essence of generating a sentiment towards a product/service is trusting the `crowd-sourcing' phenomenon, and this trust can easily be broken with deceptive content on e-commerce platforms. Such phenomena have been exposed by campaigners such as crime writer Jeremy Duns, who found a number of fellow authors involved in such practices\footnote{Amy Harmon, ``Amazon Glitch Unmasks War Of Reviewers'', New York Times, February 14, 2004.}. Consider a scenario where you are trying to launch a product as a startup, and you have established competitors in the market. If a competitor decides to hire a chain of people who post negatively about your product/service, your product may not be able to penetrate the market at all. To make matters worse, this can lead actual users to believe your product is not worth investing their money in at all. 

Identification of fake reviews has been addressed as a problem by the research community in various papers~\citep{feng2012syntactic,ott2011finding}. However, the research in this area had been limited to the identification of deceptive language~\citep{newman2003lying,mihalcea2009lie} whether they were computational methods or linguistic approaches to style detection. However,~\citet{fornaciari2014identifying} discuss a large dataset containing `real-life' examples of deceptive uses of language in a corpus they call ``DeRev'' (abbreviated from ``Deceptive Reviews''). They created this dataset in collaboration with Jeremy Duns and other `sock-puppet hunters' and showed how computational approaches to identify fake reviews can be helped with such a dataset. Similarly,~\citet{8004349} identify some indicators of fake reviews in the text based on spammer behaviour. More recently, research has attempted at identifying fake reviews in the e-commerce domain using a dataset containing text from multiple sub-domains and using an integrated neural network to help resolve the problem~\citep{alsubari2021development}. 

Apart from a social challenge like `sock-puppetry', there are always computational and linguistic challenges to any problem. For example, the domain specificity in product reviews has always existed as a `hard-to-swallow pill' for the researchers in the ABSA domain. Reviews in different domains adhere to different vocabulary and terminology, which makes it a challenge. Researchers have also attempted to find suitable cross-domain training which can help save time by training SA models in a particular domain and utilising it for another domain~\citep{sheoran-etal-2020-recommendation,schultz2018distance}. The ABSA task is a very complex endeavour in itself, and the addition of cross-domain challenges to it just makes it more challenging.~\citet{narayanan2009sentiment} has proposed that instead of focusing on a one-size-fits-all solution, researchers should focus on the many sub-problems. 

There are many critical issues when implementing an LDA-based method for ABSA~\citep{titov2008modeling}. The simple bag-of-words approach discussed in the section above also presents different challenges, and researchers have tried to use compositional semantics to help resolve some of them. Compositional semantics states that the meaning of an expression is a function of the meaning of its parts and the syntactic rules by which these are combined.~\citet{wilson2005recognizing} propose a learning algorithm that combines multiple lexicons (positive, negative, neutral) and use a compositional inference model with rules incorporated to update the deployed SVM algorithm for this task. Moreover, the language used by a genuine user of a product can be sarcastic towards the product while writing a review that seems positive at the outset. Computational approaches struggle at detecting sarcasm/irony in text, and sarcasm detection is a known challenge for researchers in the area of SA~\citep{joshi2017automatic}. Irony in the written text is a pervasive characteristic of reviews found online, and the subjectivity (or ambiguity) in the language used with online reviews is also a major challenge~\citep{reyes2012making,reyes2012humor}. Despite the challenge described above, there has been a constant push by research in the NLP and the SA domain to help come up with generic models which can help solve the challenges in this domain. We hope that this push to solve multiple sub-problems eventually leads to solving these problems. 

\section{Digital Humanities}\label{sec2.4}
Digital humanities is the confluence of computational methods (such as NLP) and areas of humanities. Digital humanities in the context of text-based datasets may analyse these datasets to understand evolution of languages, evolution of biases within the datasets. However, in the context of this chapter, we focus on approaches that use SA for digital humanities. Literary works such as books, plays and movies may be sources of entertainment, but are also reflective of the prevalent society. Analysis of literary works using computational methods can be useful to understand these literary works, compare them and analyse patterns between literary works of the same period or by the same author. For example, an interesting question may be to understand underlying common patterns and similarities between characters and plays by William Shakespeare. Similarly, another interesting application would be to analyse the Indian epics, Ramayana and Mahabharata, and compare their storylines in the context of emotion trajectories of the characters. Closer to a deployment setting, \citet{martinez2019violence} show how violence rating can be predicted using the script of a movie. Each utterance is represented as a concatenation of lexical and sentiment/violence-related features. A recurrent neural network captures the linear nature of utterances in a conversation. The final prediction is an attention-based classifier based on the neural representations of the utterances.

However, in the rest of this section, we discuss the applications of SA to applications in digital humanities. Specifically, we focus on works that focus on literature, sport and art. 

Datasets used for the purpose of emotion analysis may be provided for other areas of NLP. For example, \citet{gorinski2015movie} present a corpus of movie scripts and use it for the purpose of summarisation. Similarly, online portals such as the Internet Movie Script Database (https://www.imsdb.com/) (for movie scripts), Project Gutenberg (\url{https://www.gutenberg.org/}) (for books) may also be crawled in order to create relevant datasets.
\subsection{Applications}

Application of SA to art works can help to understand potential user experience. For example, what is the trajectory of emotions that a user will experience when reading a certain book or watching a certain movie? Understanding this may be useful to understand the impact of an art work on a person experiencing it. \citet{kim2018survey} present a detailed survey of sentiment and emotion analysis approaches applied to literary works. They highlight broad categories of approaches in the area as:
\begin{enumerate}
    \item \textbf{Classification-based formulations}: This includes tasks such as prediction of emotion of a literary work (for example, a poem), or whether or not a story has a happy ending.
    \item \textbf{Temporal sentiment changes}: This involves tasks dealing with understanding how sentiment changes over time in a given literary work. The word `time' here refers to the course of the literary work.
    \item \textbf{Relationships between characters}: This involves tasks dealing with understanding relationships between characters in a literary work. This may be interlayed with temporality where the change in relationships between characters are tracked over time.
    \item \textbf{Understanding prevalent biases}: SA applied to literary works may uncover underlying biases. For example, an emotion analysis of characters followed by marginalisation (in the probabilistic sense) over certain attributes may help to understand their portrayal. For example, it may help to understand if people of certain genders are portrayed as more angry or sad as others.
\end{enumerate}

In order to analyse emotion-related dynamics in literary works, \citet{hipson2021emotion} present a set of metrics known as utterance emotion dynamics (UED). They show how UED metrics can be calculated and can be useful in order to understand emotion dynamics in a narrative such as a book, play or a movie. Some UED metrics that can be used to analyse emotion in a literary work are:
\begin{enumerate}
    \item Emotion word density: This is the proportion of emotion words uttered by a character within a certain time interval. 
    \item Emotional arc: The emotional arc of a character is the path taken by the character over the course of the literary work. For example, the character may start off in a happy state. Then, during the course of the narrative, an unpleasant incident occurs with them, following which they become sad and angry. As the events unfold, the character becomes happy at the end of the story.
    \item Home base, variability \& displacement count: The three metrics indicate (respectively): what is the most common emotional state of a character, how much does their emotional state vary, and how often does it vary. Home base can be viewed as the most common emotional state of a character while variability indicates how much their emotional state differs in the story. Displacement count is the number of times the character experiences an emotion displacement in the story.
    \item Peak distance and rise/recovery rates: Peak distance indicates how far a character in a play deviates from their most common emotional state. Rise rate indicates how soon they achieve the peak distance while recovery rate indicates how quickly they come back to their home base. This metric potentially refers to the ups and downs in a story.
\end{enumerate}
The metrics above provide a structured framework to analyse literary works using emotion analysis.
\citet{nalisnick2013character} apply SA to analyse Shakespeare's plays. The SA approach in itself is rather simple. The sentiment values of words in an utterance are looked up in a sentiment lexicon, and the sentiment is attributed as a directed relationship between the character who spoke a certain utterance and the character who spoke the next utterance in the conversation. However, the key analysis proceeds in two dimensions: relationships between characters and time in the play. As a result, the paper describes how sentiment and affinity relationships between characters change over the course of a play. \citet{joshi2016emogram} perform a similar temporal analysis for the commentary of a cricket match. A cricket match is played over a sequence of delivery of balls. The commentary captures what happens in each of the balls. Based on what the commentator said to report each of the deliveries, the EmoGram analyses how exciting different cricket matches were.

Literary works may also provide a proxy for real life. If a dataset of real-world data is not available, creative works can serve as a useful dataset. For example, \citet{joshi2016harnessing} present an approach for sarcasm detection in conversations. Since a dataset from conversational data is not available and since they hope for a certain class distribution in the dataset, they use transcripts of the TV Show `Friends'. Every utterance in this dataset is manually annotated with sarcasm label. Then, sequence labeling algorithms are used to predict sarcasm in every utterance, using features from the utterance and utterances in its neighbourhood.

Finally, digital humanities can also use SA to create a multimedia anthology of the society. An  example of this is the `We Feel' dashboard~\citep{larsen2015we}. We Feel is an interactive dashboard maintained by the State Library of New South Wales, Australia. In order to maintain a historical anthology as seen on social media, We Feel monitors tweets, detects topics from the tweets and also records the emotion experienced by people. The We Feel dashboard contains an emotion flower - which visualises emotion changes in tweets posted from Australia and around the world. 





\section{Other Research Areas}\label{sec2.6}
There are multiple other areas of research where the analysis of sentiment can help improve the output. In this section, we delineate such areas and try to cover the known applications of SA to other research areas and the challenges it brings forth in the said research domain. Many of the research areas we cover here belong to the NLP domain but not all. As we discussed earlier, the recent SA research-based models attempt to determine the text's polarity and intensiveness. Various algorithms divide the text into different polarities and try to learn a sentiment towards a product, a topic, a policy, and so on. However, we have not yet discussed how this research applies to other areas and how these areas use automatic SA of text to perform the tasks pertaining to these areas. Let us briefly discuss such areas and see how SA applies to them.  

\subsection{Applications to Other Research Areas}

SA (SA) can benefit many research areas, including research in the non-computational domain. As deployable sentiment-based models become easier to use, researchers in various domains like psychology, interpreting studies, translation, explainability and so on have used these models to build up their research. One of the more recently investigated areas by many NLP and Deep Learning researchers is Explainable AI (XAI). XAI research has used SA as a classification task which can help explain the decision of a neural network-based model in past research~\citep{arras2017explaining,han2020explaining}. Explaining and justifying recommendations to users has the potential to increase the transparency and reliability of such models~\citep{ni2019justifying}. Existing research has tried to learn user preferences and writing styles from crowd-sourced reviews to generate explanations in the form of natural language, i.e., synthesised review generation~\citep{ni2018personalized,dong2017learning}. However, a large portion of the review text used for the SA task does not contain many features that can help identify the cause of decision-making towards either the positive or the negative class. Therefore,~\citet{ni2019justifying} extract fine-grained aspects from justifications provided in the sentiment text to come up with user personas and item profiles consisting of a set of representative aspects. They discuss two different neural network-based models that can provide justifications based on different aspects as a reference and provide justifications based on existing justification templates in the data for the sentiment text/instances that do not provide justifications. Similarly,~\citet{han2020explaining} contrast between the use of gradient-based saliency maps~\citep{li2015visualizing} and the use of Influence Functions to provide explanations for decisions made by SA models. Gradient-based saliency maps can also be visualised with the help of existing tools like LIME~\citep{ribeiro2016should}. This method computes squared partial derivatives obtained by standard gradient back-propagation. On the other hand, Layer-wise Relevance Propagation (LRP) has also been commonly used to explain the decisions of a SA model~\citep{sanchez2014eurosentiment,zucco2018explainable,ito2019concept,kaur2021proposed}. With this method, each layer neuron in the neural network is provided with a relevance score back-propagating from the final layer to the input layer neurons. However, XAI is not the only research area SA has been utilised for further research. Let us delve into some of the other past research where SA has proven to be a useful area in terms of existing research and model deployability:
\begin{description}
    \item[Psychology:]~\citet{jo2017we} argue how the sentiment expressed in a text can be used to predict the emotional state of a person and not the sentiment of a text. They build neural network-based models to perform SA and attempt to model the emotional state of a person. Based on a similar context,~\citet{10.3389/fpsyg.2019.01065} present an exploratory study on how SA can allow more empathetic automatic feedback in online cognitive behavioural therapy (iCBT) interventions for people with psychological disorders, including depression.
    \item[Translation:] The translation of sentiment is a known problem within the deep learning-based translation models. Much research has shown that perturbing the input to such translation models, \textit{i.e.,} reversing the polarity or the sentiment carried by the original input sentence results in incorrect output translation where the correct sentiment or polarity (especially negation) is not carried forward~\citep{salameh2015sentiment,mohammad2016translation,hossain2020s,saadany2020great,kanojia-etal-2021-pushing,saadany2021sentiment,tang2021revisiting} including research from statistical machine translation era~\citep{wetzel2012enriching}.
    \item[Interpreting:]~\citet{carstensen2017court} argue how interpreting during a court session, \textit{i.e.,} legal interpreting required the use of emotions by human interpreters to make it better. As automated methods for interpreting become more used in this area, such models will need to take the `sentiment' expressed by various parties into stock. Similarly, interpreting in the healthcare domain can lead to situations of distress among both the interpreter~\citep{korpal2019investigating} and the person describing their illness~\citep{tiselius2020distressful}. The person describing their illness can modify their emotions if they feel distressed describing their ailment to the medical professional, which is being interpreted due to a language barrier~\citep{OUT2020101255,ValeroGarcs2016EmotionalAP}.
    \item[Sociology:]~\citet{karamibekr2012sentiment} discuss the use of SA for social issues. They discuss how SA of social issues can affect social policy and the political scenario of a country. They present their views on how SA is not only important to the product review domain, and the applications of SA research should not be limited to the e-commerce domain. A more descriptive discussion on how SA is essential for social policy can be found in Section~\ref{sec2.2} in this chapter. 
\end{description}

With the aforementioned applications of SA research in various domains, we further discuss some challenges towards applying sentiment research in these areas. 

\subsection{Challenges for SA Research in Other Areas}

Applying SA research in other domains comes ubiquitously for NLP research domains like Translation, Question Answering, Conversational Agents and so on. However, it is incredibly challenging if SA research is used for the XAI domain. Visualisation of neural models in NLP is a challenge because of the following factors~\citep{li2015visualizing}:
\begin{description}
    \item[Compositionality:] It is not clear how vector-based models achieve `compositionality' \textit{i.e.,} building sentence vector/meaning from the vector/meaning of words and phrases.
    \item[Interpretability:] The `interpretability' and `causality' of the decision-making is questioned on many occasions making them less `reliable' for real-world applications like decisions on parole applications\footnote{\href{https://www.technologyreview.com/2019/01/21/137783/algorithms-criminal-justice-ai/}{MIT Technology Review: Criminal Justice}} and many more.  
    \item[Complexity:] Deep learning models have become increasingly complex, and unfortunately, their inscrutability has grown in tandem with their predictive power~\citep{doshivelez2017rigorous}. As multi-layered neural models (\textit{e.g.,} BiLSTM, Multimodal models) are deployed for solving problems in the AI domain; they become more and more complex to explain. 
    \item[Hidden Layers:] The very black-box nature of neural models is due to the existence of hidden layers in the neural architectures constantly used for solving NLP problems/tasks.
\end{description}

Researchers have argued upon the very definition of \textit{interpretability} or \textit{causality} for neural models. \textit{Interpretability} is used to confirm other important desiderata of ML/DL-based models or NLP systems. There exist many auxiliary criteria like the notion of \textit{fairness} or \textit{unbiasedness} towards protected groups which needs further investigation to ensure they are not discriminated against by other groups. \textit{Causality} is sometimes defined as the predicted change in output due to a perturbation that may occur in the input for a real-world NLP system, and SA research has always played a prominent role in the development of important XAI concepts which are now fundamental to the NLP research in the XAI domain. Some areas such as \textit{fairness} and \textit{privacy} have formalised the criteria for causality in their research. These formalisations have allowed for more rigorous research in these fields. In many areas such as psychology, however, such formalisation remain elusive, and arguments are presented on the lines of ``explanations may highlight an incompleteness''~\citep{keil2004lies}. But research from the SA domain combined with XAI can assist in qualitatively ascertaining whether other desiderata such as fairness, reliability, transparency and so on, are met. Given that much SA research is performed on limited datasets, there are many challenges towards actually using these models in a real-world scenario but nonetheless, this research can be used as a proof-of-concept to build more ``mature'' DL-based models which can help solve real-world problems. In the next section, we summarise this chapter and briefly discuss the research presented above. 

\section{Summary}\label{sec2.7}
Applications of sentiment analysis (SA) to real-life scenarios are the last mile for SA: they make the research in SA useful for specific real-world tasks. In this chapter, we describe five applications of SA. The first one is health. In terms of health, we describe applications such as social media-based epidemic intelligence, while highlighting challenges around selection bias. The next application is social policy, wherein we describe how SA can be used that understand public sentiment about social issues while highlighting issues such as noisy data and privacy. The next application is e-commerce which focuses on the use of SA for business objectives. This area in particular witnesses a breadth of SA techniques for specific business problems such as aspect-based SA. It is also important, however, to note that fake reviews are an issue that such an application must deal with. The next application of SA is digital humanities where we describe approaches that use SA on literary works and other cultural artifacts for an improved understanding of the artwork. Finally, we also describe how SA has impacted other areas of NLP such as translation and interpretation, requiring a focus on compositionality and interpretability that are crucial for both SA and the related NLP task.

As SA continues to be deployed in several applications impacting businesses, individuals and the society, the applications described in this chapter provide an introductory perspective to the breadth of SA and the impact SA has had on several real-life scenarios.


%
%
%
%
%
%
%
%
%
\bibliographystyle{elsarticle-harv}
%
%
%
%
%
%
%
%
%
%
%
\bibliography{book.bib}
%
%
%
%
%
%
\end{document}